\documentclass[journal=jacsat,manuscript=article]{achemso}

\usepackage[version=3]{mhchem} 
\usepackage{float}
\usepackage{amssymb}
\usepackage{amsmath}
\usepackage{algorithm}
\usepackage{verbatim}
\usepackage[makeroom]{cancel}
\makeatletter
\newcommand*{\addFileDependency}[1]{
  \typeout{(#1)}
  \@addtofilelist{#1}
  \IfFileExists{#1}{}{\typeout{No file #1.}}
}
\makeatother

\usepackage{hyperref}
\hypersetup{
    colorlinks=true,
    linkcolor=blue,
    citecolor=blue,
    filecolor=magenta,      
    urlcolor=blue,
    pdfpagemode=FullScreen,
    }

\author{Wei-Tse Hsu}
\affiliation{Department of Biochemistry, University of Oxford, South Parks Road, Oxford, OX1 3QU, UK}

\author{Savva Grevtsev}
\affiliation{Department of Chemistry, University of Oxford, Mansfield Road, Oxford, OX1 3TA, UK}

\author{Thomas Douglas}
\affiliation{Department of Biochemistry, University of Oxford, South Parks Road, Oxford, OX1 3QU, UK}

\author{Aniket Magarkar}
\affiliation{Boehringer Ingelheim Pharma GmbH \& Co. KG, Birkendorfer Str. 65, 88397 Biberach an der Riß, Germany}
\email{aniket.magarkar@boehringer-ingelheim.com}

\author{Philip C. Biggin}
\affiliation{Department of Biochemistry, University of Oxford, South Parks Road, Oxford, OX1 3QU, UK}
\email{philip.biggin@bioch.ox.ac.uk}

\title
  {Can AI-predicted complexes teach machine learning to compute drug binding affinity?}


\begin{document}



\clearpage
\begin{abstract}
We evaluate the feasibility of using co-folding models for synthetic data augmentation in training machine learning-based scoring functions (MLSFs) for binding affinity prediction. Our results show that performance gains depend critically on the structural quality of augmented data. In light of this, we established simple heuristics for identifying high-quality co-folding predictions without reference structures, enabling them to substitute for experimental structures in MLSF training. Our study informs future data augmentation strategies based on co-folding models.
\end{abstract}

\clearpage
\section{Introduction}
Over the past decades, machine learning-based scoring functions (MLSFs) have gained increasing popularity in computer-aided drug discovery~\cite{sellwood2018artificial}. By leveraging 3D structures of binding complexes--commonly protein-ligand binding complexes--these models predict binding affinities in a fraction of the time required by physics-based simulation methods such as alchemical free energy perturbation~\cite{king2021recent}, while achieving arguably comparable accuracy in some scenarios~\cite{valsson2025narrowing, passaro2025boltz}. During training, they often rely on experimental structures of binding complexes, representing binding interfaces with underlying architectures ranging from feed-forward neural networks~\cite{meli2021learning}, convolutional neural networks~\cite{wang2021onionnet}, to graph neural networks~\cite{valsson2025narrowing, mqawass2024graphlambda}. However, the data of high-resolution experimental complexes with matched binding affinity measurements remain rare, limiting both the scale and diversity of training datasets available for these models.

To address this scarcity, several efforts have emerged to synthetically augment training datasets using computational modelling. One notable example is BindingNet~\cite{li2024high}, which uses protein structures from PDBbind~\cite{wang2005pdbbind} as templates and models new complexes by aligning structurally similar ChEMBL~\cite{gaulton2012chembl} ligands to the reference ligands based on their maximum common substructures. With this template-based modelling approach, BindingNet v1 generated approximately 70K protein-ligand complexes with associated activity data from ChEMBL. Its successor, BindingNet v2~\cite{zhu2025augmented}, introduced a hierarchical variation of the modelling pipeline to accommodate less similar candidate ligands, further expanding the dataset to roughly 700K complexes.  Recent studies have demonstrated that the inclusion of BindingNet v1 improves the performance of MLSFs~\cite{valsson2025narrowing}, though BindingNet v2 has so far only been used to train the docking model Uni-Mol~\cite{zhou2023unimol}, where improved success rates in PoseBusters~\cite{buttenschoen2024posebusters} sanity checks were observed. 

One inherent drawback of these template-based modelling approaches, however, is their reliance on high-quality, experimentally determined protein structures as templates, which restricts the extent of data augmentation. Moreover, these methods implicitly assume that structurally similar ligands that bind to the same protein receptor share the same binding mode, an oversimplification that does not always hold in practice. Recent advances in co-folding models, such as AlphaFold3 (AF3)~\cite{abramson2024accurate}, Chai-1~\cite{chai2024chai}, and Boltz-1~\cite{wohlwend2024boltz}/Boltz-2\cite{passaro2025boltz}, enable \textit{de novo} structure prediction of protein-ligand complex structures, offering a promising alternative to further expand the scope of binding complex datasets. Indeed, a large-scale dataset generated using Boltz-1 has been recently proposed by Lemos et al.~\cite{lemos2025sair} Yet, the use of co-folding predictions for large-scale dataset generation has not been systematically examined in the context of training MLSFs. 

In this work, we report key insights from training AEV-PLIG~\cite{valsson2025narrowing}, a state-of-the-art GNN-based scoring function, on multiple modelled complex datasets, including BindingNet v1, BindingNet v2, and Boltz-1x-based reproductions of a recently introduced experimental dataset HiQBind~\cite{wang2025workflow}. Our study focuses on three fundamental questions: (1) To what extent does data augmentation improve MLSF performance, especially when introduced with synthetic training examples that are not necessarily of high quality? (2) Are co-folding predictions sufficient to replace or complement experimental structures for MLSF training? (3) What practical heuristics can be used to identify high-quality co-folding predictions in the absence of reference structures? By systematically addressing these questions, we aim to provide early but essential insights into how best to leverage co-folding models in large-scale dataset construction for structure-based machine learning. 


\section{Results and Discussion}
\subsection{Data augmentation benefits can be diluted by low-quality examples}
\begin{figure}[H]
    \centering
    \includegraphics[width=\textwidth]{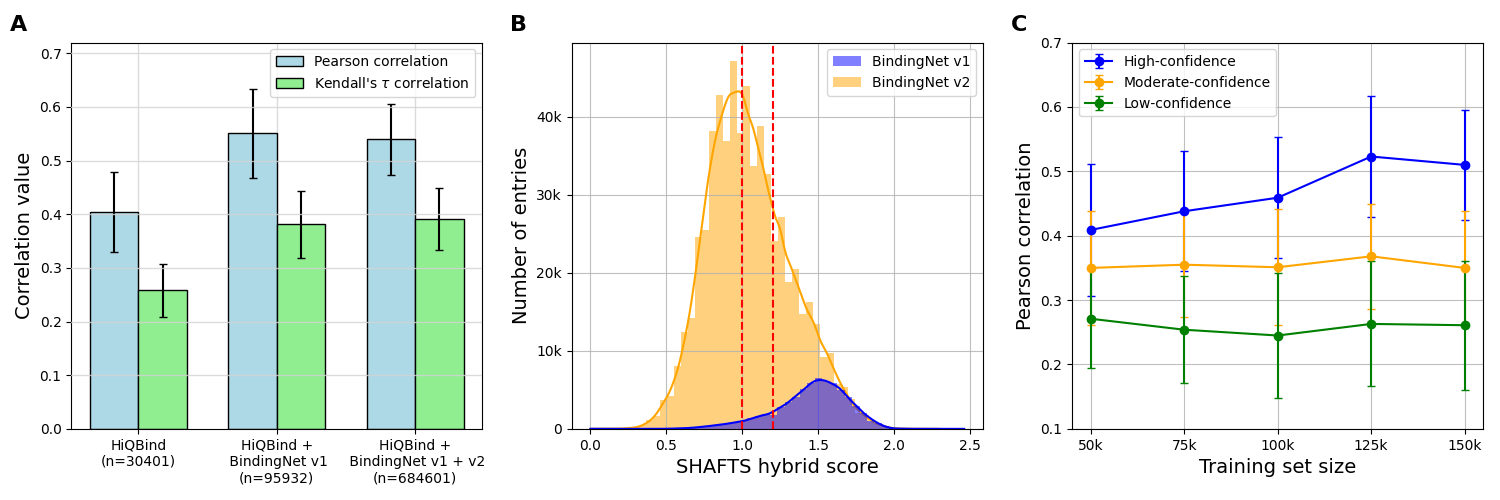}   
    \caption{Performance of AEV-PLIGs trained on various combinations and partitions of HiQBind, BindingNet v1, and BindingNet v2. (A) Model performance when trained on HiQBind alone, HiQBind + BindingNet v1, and HiQBind + BindingNet v1 + BindingNet v2. The sizes of the datasets are noted in the parentheses of the labels. (B) Distribution of SHAFTS hybrid scores in BindingNet v1 and v2, with two vertical lines marking the cutoffs for different confidence partitions. (C) Performance of models trained on progressively larger subsets of BindingNet v1 + v2, constructed from different confidence partitions. Each larger subset includes all smaller ones.}
    \label{data_augmentation_result}
\end{figure}

We first assessed the impact of synthetic data augmentation on the performance of machine learning-based scoring functions. Figure ~\ref{data_augmentation_result}A presents the performance of AEV-PLIGs trained on different combinations of HiQBind, BindingNet v1, and v2. Performance was evaluated using Pearson correlation coefficient (PCC) for scoring power and Kendall's $\tau$ for ranking power, both computed between the predicted and experimentally measured binding affinities on the FEP benchmark dataset~\cite{ross2023maximal}, a challenging test set with minimal data leakage from our training sets (see Figure S1). As a result, the addition of BindingNet v1 substantially improved model performance, consistent with previous findings~\cite{valsson2025narrowing}. Interestingly, the inclusion of BindingNet v2--despite an increase in the training set size by over sevenfold--did not yield any further noticeable improvement. 

To gain more insights into this, we examined the SHAFTS~\cite{liu2011shafts} hybrid score, a confidence metric used in BindingNet. According to the original BindingNet v2 study~\cite{zhu2025augmented}, structures are categorised as high-confidence if their hybrid score exceeds 1.2, moderate-confidence if between 1.0 and 1.2, and low-confidence if below 1.0. As shown in Figure~\ref{data_augmentation_result}B, most entries in BindingNet v2 fall into the low- to moderate-confidence range, whereas BindingNet v1 exhibits a markedly higher median confidence score (1.48 versus 1.02). The original study~\cite{zhu2025augmented} reported a top-1 docking success rate (defined as a ligand RMSD $<$ 2 \r{A}) of only 16\% for low-confidence structures, 33\% for moderate-confidence, and 73\% for high-confidence entries. The overrepresentation of lower-confidence entries in BindingNet v2 therefore strongly suggests that it contains a significantly higher proportion of low-quality structures than v1.

To quantify how such differences in data quality affect model performance, we trained AEV-PLIGs on subsets of the union of BindingNet v1 and v2 grouped by confidence levels. For each confidence partition, these subsets were constructed by successively introducing entries randomly drawn from the partition (Figure \ref{data_augmentation_result}C). As a result, when models were trained solely on high-confidence structures, the model performance generally improved with increasing training set size, although some suboptimal structures in the high-confidence subset may have tempered the overall gains. In contrast, no such improvement was observed for models trained on moderate- and low-confidence data. This is reflected by the Kendall's $\tau$ correlation between PCC and training set size, which was 0.80 for high-confidence, but only 0.105 and $-$0.20 for moderate- and low-confidence subsets, respectively. Together, the results shown in Figures \ref{data_augmentation_result}B and C help explain the negligible performance change upon the inclusion of BindingNet v2 in the training set. These findings demonstrate that data augmentation is effective for MLSF training only when the structural quality of newly introduced examples is sufficiently high. Simply adding more synthetic complexes--without quality control--offers limited benefit, highlighting the need for rigorous filtering in future dataset construction efforts.



\subsection{Practical heuristics support reliable curation of co-folding predictions}
As the utility of synthetic data in MLSF training hinges on structural quality, one key challenge in data augmentation using co-folding models is to identify reliable predictions in the absence of reference structures. We therefore investigated whether simple heuristics could serve this purpose by analysing Boltz-1x predictions on a series of structure reproduction tasks. 

\begin{figure}[H]
    \centering
    \includegraphics[width=\textwidth]{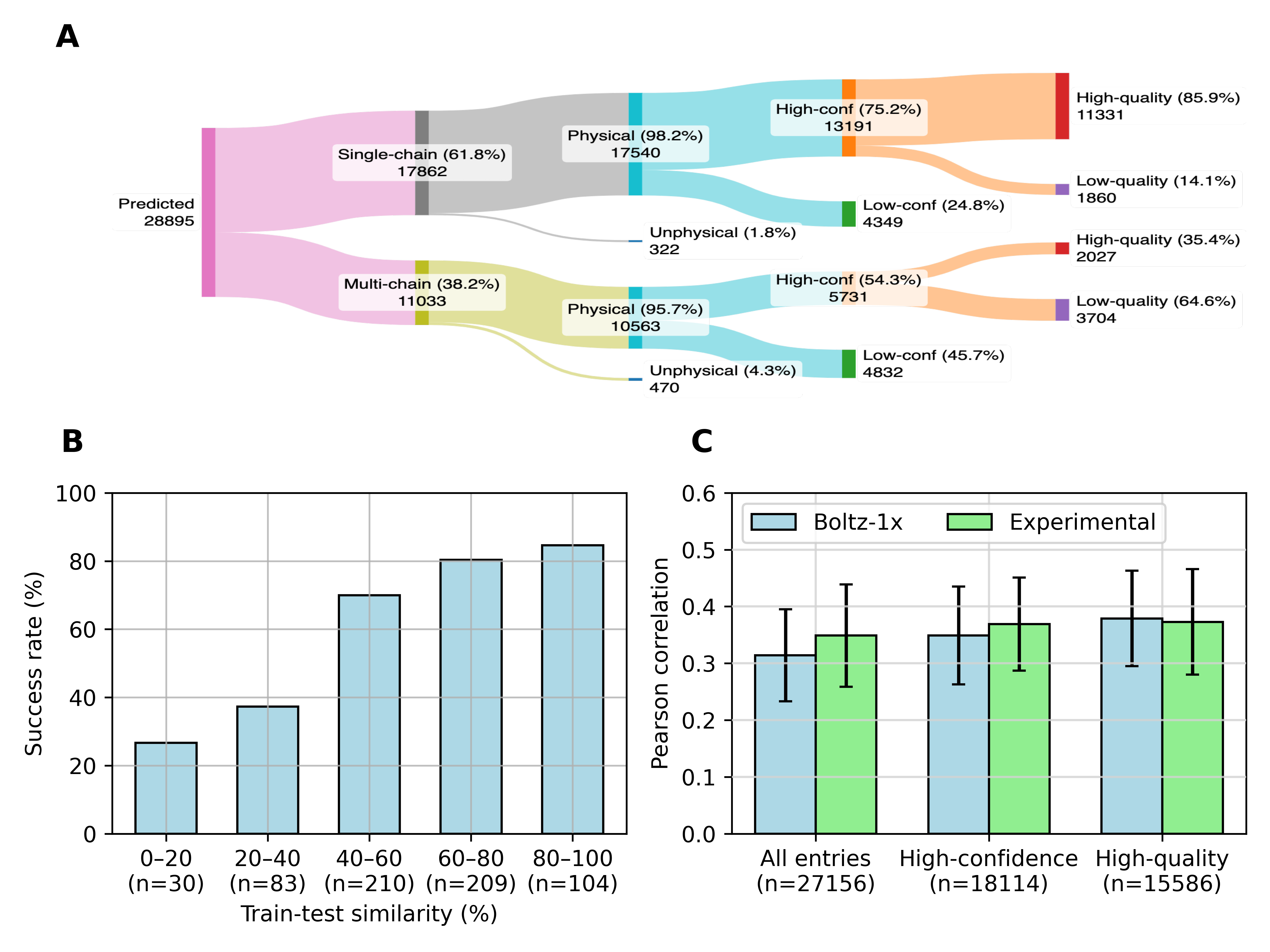}   
    \caption{Results based on the HiQBind and RNP reproduction tasks. (A) Sankey diagram that summarises the overall performance of Boltz-1x in the HiQBind reproduction task. Each flow is annotated with the number of predicted structures and its percentage relative to the preceding category. (B) Success rate (defined as the percentage of structures having a pocket RMSD $<$ 2 \r{A}) in the RNP reproduction task given different levels of train-test similarity defined in the RNP study,~\cite{vskrinjar2025have} which is a product of binding pocket coverage~\cite{durairaj2024plinder} and combined overlap score (SuCOS)~\cite{malhotra2017does} of the ligand pose. (C) Performance of AEV-PLIGs trained on different subsets of HiQBind and their Boltz-1x reproduced counterpart. The size of the training set in each case is annotated. Details about training set curation are available in the Methods section in the Supporting Information.} 
    \label{reproduction}
\end{figure}

As a starting point, we used Boltz-1x~\cite{passaro2025boltz} to reproduce the recently introduced HiQBind dataset~\cite{wang2025workflow}, which is arguably the highest-quality experimental dataset of protein-ligand complexes available for MLSF training. Boltz-1x has been shown to be among the strongest co-folding models for protein-ligand structure prediction, particularly in generating physically plausible structures~\cite{nittinger2025co, jiang2025posex}, making it an ideal choice for our task. Specifically, we replaced each experimental structure in HiQBind with its corresponding Boltz-1x prediction, then compared the prediction with the experimental reference to assess its quality. As illustrated in the Sankey diagram in Figure \ref{reproduction}A, Boltz-1x exhibited greater confidence when predicting complexes with single-chain receptors: among structures passing the PoseBusters~\cite{buttenschoen2024posebusters} sanity check (labelled ``Physical''), 75.2\% had a Boltz confidence score above 0.9 (labelled ``High-conf''). Although the Kendall's $\tau$ correlation between commonly used confidence and quality metrics is generally weak (Figure S2)--indicating that higher confidence does not necessarily imply higher structural quality--for single-chain systems, a simple confidence threshold of 0.9 still usefully identifies a subset in which 85.9\% of predictions are high-quality (pocket RMSD $<$ 2 \r{A}).

Note, however, that HiQBind is within the training set of Boltz-1, which explains the decent performance observed in the reproduction task. To examine whether similar filtering heuristics remain useful for unseen data, we used Boltz-1x to reproduce Runs and Poses (RNP) proposed by Škrinjar et al.~\cite{vskrinjar2025have}, a dataset comprising 2600 binding complex structures released after the training cutoff date of AF3-like co-folding models and specifically designed to evaluate their generalisability. We focused on single-chain structures, which, as shown in Figure \ref{reproduction}A, are more reliably predicted by Boltz-1x and represent the most practical targets for co-folding data augmentation. A Sankey diagram summarising the basic statistics for the RNP reproduction task is shown in Figure S3.

As a result, Figure \ref{reproduction}B shows that the success rate (defined as the percentage of predicted complexes having a pocket RMSD $<$ 2 \r{A}) declines with the train-test similarity, which agrees with the findings in the work by Škrinjar et al.~\cite{vskrinjar2025have} Nonetheless, predictions with 60\% to 80\% train-test similarity still achieved a success rate of approximately 80\%, suggesting that predictions above this threshold remain reasonably reliable. Given the architectural similarities and comparable performance reported across AF3-derived co-folding models~\cite{wohlwend2024boltz, vskrinjar2025have}, we expect this threshold to generalise. Overall, these findings support the feasibility of applying simple heuristics to guide reliable co-folding-based dataset construction at scale.

\subsection{Co-folded structures suffice for MLSF training}
To assess whether co-folding predictions can serve as a substitute for experimental structures in MLSF training, we compared AEV-PLIGs trained on different subsets of HiQBind and their Boltz-1x-reproduced counterparts. As shown in Figure \ref{reproduction}C, AEV-PLIGs trained on Boltz-1x predictions achieved performance statistically indistinguishable from those trained on the original experimental structures--whether using the full dataset, only high-confidence predictions (Boltz-1x confidence score $>$ 0.9), or only high-quality predictions (pocket RMSD $<$ 2 \r{A} and validated by PoseBusters). This suggests that co-folded structures can provide training signal nearly equivalent to that of experimental structures. Notably, including all Boltz-1x predictions, which led to a training set 74\% larger than the high-quality subset, did not lead to improved performance. This again mirrors the observations in Figure ~\ref{data_augmentation_result}A, reinforcing that data augmentation with low-quality examples offers little benefit in improving scoring function performance. 


\section{Conclusions}
High-quality binding structures are critical for training effective machine learning-based scoring functions, yet their scarcity remains a limiting factor. This study demonstrates that co-folding predictions, when properly filtered, can serve as viable substitutes for experimental structures in large-scale MLSF training, offering comparable model performance even when used as full replacements. 

Through systematic evaluation, we established when synthetic augmentation improves model performance, how to reliably select useful co-folding predictions, and that filtered co-folded structures can match experimental ones in MLSF training. In particular, we found that low-quality synthetic examples offer little benefit even when they dramatically increase training set size, underscoring the need for stringent quality control in data augmentation. We further established practical heuristics, such as prioritising single-chain complexes, filtering by Boltz-1x confidence score $>$ 0.9, and enforcing a train-test similarity above 60\%, to effectively identify high-quality predictions in the absence of reference structures. These filtering strategies enable co-folding predictions to be used at scale without compromising model accuracy.

Taken together, our findings provide a practical foundation for extending MLSF datasets beyond experimentally determined structures, particularly in underrepresented protein families where structural data remain sparse. By enabling scalable, quality-controlled data augmentation, co-folding models hold promise for advancing the next generation of structure-based machine learning in drug discovery.

\section{Code availability}
All AEV-PLIG experiments in this study were conducted using a refined version of the AEV-PLIG codebase, available under the 3-Clause BSD License: \href{https://github.com/wehs7661/AEV-PLIG-refined}{https://github.com/wehs7661/AEV-PLIG-refined}. Large-scale Boltz-1x predictions and subsequent analyses were performed using in-house code, which will be publicly released soon.

\section*{Author Contributions}
W.-T.H. primarily conceptualised the project, with contributions from T.D., A.M. and P.C.B. W.-T.H. performed the experiments, with contributions from S.G. in refining relevant codebases and T.D. in preliminary testing. W.-T.H. wrote the original manuscript draft. A.M. and P.C.B. edited and reviewed the manuscript. A.M., and P.C.B. supervised the project and obtained resources.

\section{Acknowledgements}
We gratefully acknowledge support from the DAWN AI Research Resource, which is funded by UK Research and Innovation (UKRI) as part of the AIRR Early Access Project )ANON-BYYG-VXG7-7).




 

\clearpage
\providecommand{\latin}[1]{#1}
\makeatletter
\providecommand{\doi}
  {\begingroup\let\do\@makeother\dospecials
  \catcode`\{=1 \catcode`\}=2 \doi@aux}
\providecommand{\doi@aux}[1]{\endgroup\texttt{#1}}
\makeatother
\providecommand*\mcitethebibliography{\thebibliography}
\csname @ifundefined\endcsname{endmcitethebibliography}  {\let\endmcitethebibliography\endthebibliography}{}


\end{document}



\clearpage
\section{Methods}
\subsection{AEV-PLIG training}
For all AEV-PLIG experiments in this study, we trained an ensemble of five models using the default hyperparameters from the original AEV-PLIG implementation~\cite{valsson2025narrowing} using NVIDIA A40 GPUs. Specifically, we used a batch size of 128, 200 training epochs, 256 hidden dimensions, 3 attention heads, a learning rate of 0.000123, and leaky ReLU~\cite{maas2013rectifier} as the activation function. Complexes that could not be processed by RDKit~\cite{landrum2016rdkit} were discarded, and ligands having uncommon elements (i.e., elements other than H, B, C, N, O, F, P, S, Cl, Br, or I) were excluded. To prevent data leakage, any complex whose ligands exhibit a Tanimoto similarity greater than 0.9 to any ligand in the test set (the FEP benchmark) was excluded from the training set. It is worth noting that this choice of similarity cutoff should have a negligible impact on AEV-PLIG training, as the FEP benchmark exhibits minimal chemical overlap with the training sets used. As shown in Figure S1, the vast majority of ligands in HiQBind, BindingNet v1, and BindingNet v2 have low maximum Tanimoto similarity to any ligand in the FEP benchmark.

\subsection{AEV-PLIG testing}
All trained scoring functions in this study were tested using the FEP benchmark curated by Ross et al.~\cite{ross2023maximal}, which comprises congeneric series commonly seen in real-world drug discovery projects. The dataset covers a wide variety of protein targets and ligands and has minimal overlap with the training sets used in this study, including HiQBind, BindingNet v1, and BindingNet v2. Since most machine learning-based scoring functions (MLSFs) are trained to predict the negative logarithm of binding affinity ($\text{pK}$), we converted the experimental binding free energy values ($\Delta G$) in the FEP benchmark to $\text{pK}$ values using the following equation $$\Delta G = -\ln(10)RT\text{pK}$$ where the gas constant $R=1.987 \times 10^{-3}$ $\mathrm{ kcal \cdot K^{-1} \cdot mol^{-1}}$ and the temperature $T=297\mathrm{K}$. 

We then respectively used the Pearson correlation coefficient and Kendall's $\tau$ correlation coefficient to assess the scoring power and ranking power of each trained scoring function. All correlation values reported in the main text are weighted averages across FEP benchmark series having ten or more ligands. Uncertainties are reported as bootstrapped 95\% confidence intervals. 

\subsection{Reproduction tasks using Boltz-1x}
\subsubsection{HiQBind reproduction}
HiQBind, recently proposed by Wang et al.~\cite{wang2025workflow}, contains 32275 protein-ligand complexes curated from Binding MOAD~\cite{hu2005binding}, BindingDB~\cite{liu2007bindingdb}, and BioLiP~\cite{yang2012biolip}, with structural artifacts corrected using a semi-automated computational workflow. For Boltz-1x reproduction, we excluded 3469 entries containing non-standard amino acids in the receptor and 82 entries with conformer generation issues from RDKit~\cite{landrum2016rdkit} for the ligand. Additional exclusions were made for entries that failed due to out-of-memory issues during inference or could not be processed by our automatic analysis workflow due to malformed PDB references. The final reproduced dataset contains 28895 predicted structures. 

\subsubsection{RNP reproduction}
Runs and Poses (RNP), recently proposed by Škrinjar et al.~\cite{vskrinjar2025have}, is a benchmark of 2600 experimentally determined protein-ligand complex structures released after the training cutoff date of AF3-like co-folding models. For Boltz-1x reproduction, we applied the same filtering criteria used for HiQBind, with an additional restriction to single-chain receptor entries. This ultimately led to a final set of 636 reproduced structures for downstream analysis. 

\subsubsection{Common settings}
Both reproduction tasks were performed using NVIDIA A40 GPUs with default settings for Boltz-1x inference, which include 3 recycling steps, 200 sampling steps, a diffusion step size of 1.638, with a greedy MSA pairing strategy. MSA computation was done using the MSA server provided by ColabFold~\cite{mirdita2022colabfold}. For entries sharing the same receptor sequence, we reused the same MSA results to accelerate the prediction pipeline. For each binding complex, only one diffusion sample was generated. 

\subsubsection{Analysis}
Boltz-1x provides a wide range of confidence metrics, including pTM, ipTM, ligand ipTM, protein ipTM, complex pLDDT, complex PDE, complex iPDE, among others. Definitions of these metrics are available in the original Boltz-1x technical report.~\cite{wohlwend2024boltz} In addition, we compute the DOPE (Discrete Optimised Protein Energy) score~\cite{shen2006statistical} to assess the energetic plausibility of the predicted protein structure. Region-specific pLDDT scores are also calculated from the pLDDT matrices provided by Boltz-1x inference, including ligand pLDDT, pocket pLDDT, and shell pLDDT, where pocket and shell residues are defined as those within 0-6 \r{A} and 6-8 \r{A}, respectively, of any ligand heavy atom. 

To assess the structural quality of Boltz-1x predictions relative to experimental references, we compute complex RMSD, protein RMSD, ligand RMSD, and pocket RMSD, considering only heavy atoms. For pocket RMSD, the binding pocket is defined based on the reference structure as all residues with any atom within 6 \r{A} of any ligand atom. Structures are aligned using the pocket residues from the reference, and pocket RMSD is then computed over both the pocket residues and the ligand. In our study, a structure is defined as high-quality if its pocket RMSD is below 2 \r{A}.

\section{Supplementary Figures}
\begin{figure}[H]
    \centering
    \includegraphics[width=\textwidth]{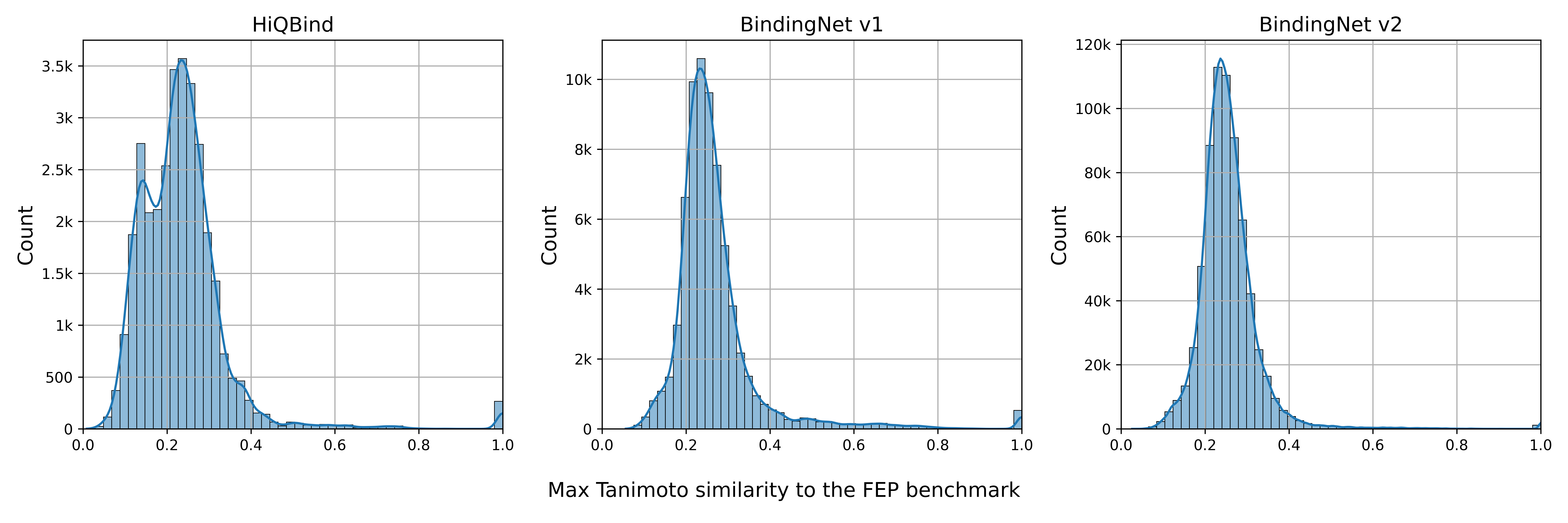}   
    \caption{Distribution of maximum Tanimoto similarity between ligands in the FEP benchmark and those in the training sets used in this study: (A) HiQBind, (B) BindingNet v1, and (C) BindingNet v2. All three training sets exhibit minimal ligand overlap with the FEP benchmark, with 97.9\%, 96.2\%, and 99.1\% of entries, respectively, having a maximum similarity below 0.5.} 
\end{figure}


\begin{figure}[H]
    \centering
    \includegraphics[width=\textwidth]{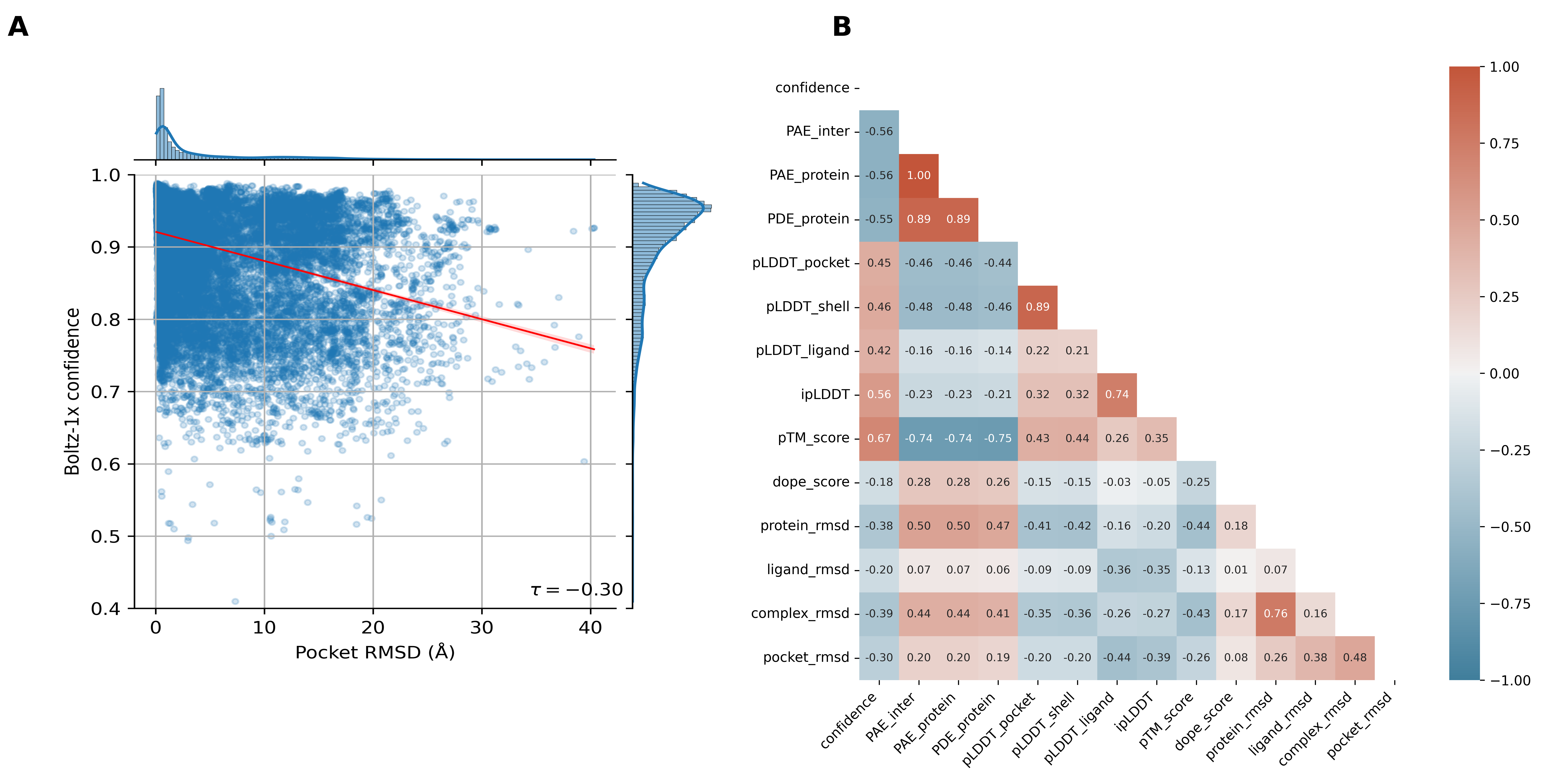}   
    \caption{Correlation analysis on Boltz-1x reproduced HiQBind. (A) Scatter plot of Boltz-1x confidence score versus pocket RMSD as a representative example, with the Kendall's $\tau$ correlation coefficient annotated. (B) Pairwise Kendall's $\tau$ correlation coefficients between commonly used confidence metrics and quality metrics. Both panels show that there is a generally weak confidence-quality correlation in Boltz-1x predictions in the HiQBind reproduction task. More details about how these metrics are defined can be found in the Methods section.} 
\end{figure}


\begin{figure}[H]
    \centering
    \includegraphics[width=\textwidth]{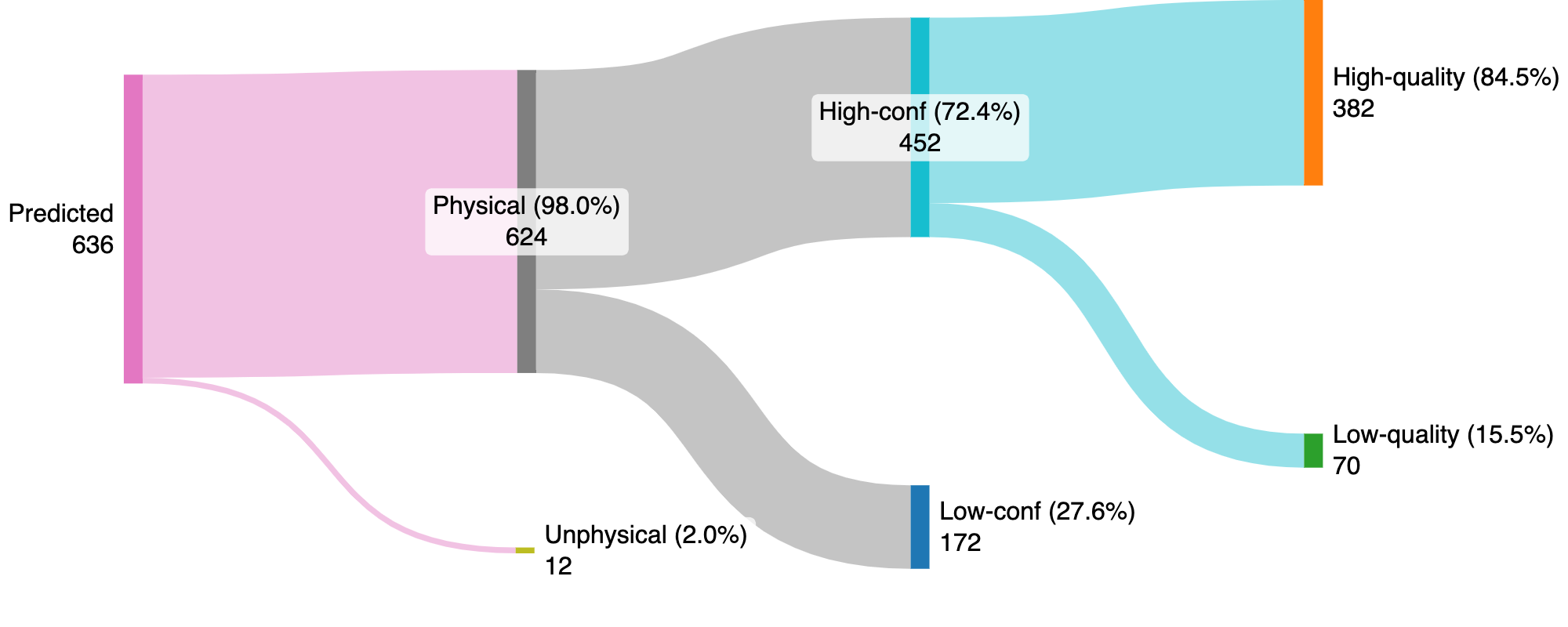}   
    \caption{Sankey diagram that summarises the overall performance of Boltz-1x in the RNP reproduction task. Each flow is annotated with the number of predicted structures and its percentage relative to the preceding category. Overall, the proportions of high-confidence predictions and high-quality structures among them are comparable to those observed for single-chain complexes in the HiQBind reproduction task.} 
\end{figure}



 

\clearpage
\providecommand{\latin}[1]{#1}
\makeatletter
\providecommand{\doi}
  {\begingroup\let\do\@makeother\dospecials
  \catcode`\{=1 \catcode`\}=2 \doi@aux}
\providecommand{\doi@aux}[1]{\endgroup\texttt{#1}}
\makeatother
\providecommand*\mcitethebibliography{\thebibliography}
\csname @ifundefined\endcsname{endmcitethebibliography}  {\let\endmcitethebibliography\endthebibliography}{}
